\documentclass{article}

\usepackage[final]{corl_2020} 

\usepackage{times}
\usepackage{epsfig}
\usepackage{graphicx}
\usepackage{amsmath}
\usepackage{amssymb}
\usepackage{booktabs ,float}

\usepackage{amssymb}
\usepackage{subfig}
\usepackage{multicol}
\usepackage{graphicx,color,subfig}
\usepackage{amsfonts, amssymb, amsmath,verbatim, multirow}
\usepackage{enumerate}
\usepackage{breqn}
\usepackage{amsmath}
\usepackage{enumitem}
\usepackage{times}
\usepackage{algorithm}
\usepackage{algorithmic}
\usepackage{booktabs}
\usepackage[numbers]{natbib}
\usepackage{multicol}
\usepackage{wrapfig}
\usepackage{appendix}

\newtheorem{theorem}{\textbf{Theorem}}[section]

\newtheorem{definition}{\textbf{Definition}}[section]

\DeclareMathOperator*{\argmax}{arg\,max}

\newcommand{\mb}[1]{{\boldsymbol{#1}}}

\newcommand{\trsp}{{\!\scriptscriptstyle\top}}

\newcommand\scalemath[2]{\scalebox{#1}{\mbox{\ensuremath{\displaystyle #2}}}}
\newcommand{\ajay}[1]{\textcolor{blue}{#1}}

\title{DIRL: Domain-Invariant Representation Learning for Sim-to-Real Transfer}

\author{
  Ajay Kumar Tanwani \\
  University of California, Berkeley \\
  \texttt{ajay.tanwani@berkeley.edu} \\
}

\begin{document}
\maketitle

\begin{abstract}
Generating large-scale synthetic data in simulation is a feasible alternative to collecting/labelling real data for training vision-based deep learning models, albeit the modelling inaccuracies do not generalize to the physical world. In this paper, we present a domain-invariant representation learning (DIRL) algorithm to adapt deep models to the physical environment with a small amount of real data. Existing approaches that only mitigate the covariate shift by aligning the marginal distributions across the domains and assume the conditional distributions to be domain-invariant can lead to ambiguous transfer in real scenarios. We propose to jointly align the marginal (input domains) and the conditional (output labels) distributions to mitigate the covariate and the conditional shift across the domains with adversarial learning, and combine it with a triplet distribution loss to make the conditional distributions disjoint in the shared feature space. Experiments on digit domains yield state-of-the-art performance on challenging benchmarks, while sim-to-real transfer of object recognition for vision-based decluttering with a mobile robot improves from $26.8 \%$ to $91.0 \%$, resulting in $86.5 \%$ grasping accuracy of a wide variety of objects. Code and supplementary details are available at: \ajay{\url{https://sites.google.com/view/dirl}}
\end{abstract}



\section{Introduction}
	
Data-driven deep learning models tend to perform well when plenty of labelled training data is available and the testing data is drawn from the same distribution as that of the training data. Collecting and labelling large scale domain-specific training data for robotics applications, however, is time consuming and cumbersome~\citep{Levine_armfarm_18}. Additionally, the sample selection bias in data collection limits the model use to very specific environmental situations. Training deep models in simulation for robot manipulation is becoming a popular alternative~\citep{Tobin_sim2real_17,Peng_sim2real_17,openai_2019}. Despite the efforts to build good dynamic models and high fidelity simulators for scalable data collection, the modeling inaccuracies makes it difficult to transfer the desired behaviour in real robot environments. 

This paper investigates the problem of sample-efficient domain adaptation to learn a deep model that can be transferred to a new domain~\citep{Wang_da_survey_18}. We analyze the situation where we have a lot of labeled training data from the simulator or the \textit{source} distribution, and we are interested in adapting the model with limited or no labelled training data drawn from the \textit{target} distribution. Existing popular approaches to domain adaptation learn common feature transformations by aligning marginal distributions across domains in an unsupervised manner~\citep{Ganin_DANN_16,Tzeng_adda_17,Hoffman_cycada_17,Saito_MCDDA_18}. They implicitly assume that the class conditional distributions of the transformed features is also similar across domains, i.e., if the model performs well on the source domain and the features overlap across domains, the model is going to perform well on the target domain. This, however, creates ambiguity in class alignment across source and target domains in the shared feature space that can result in a negative transfer to a new domain as shown in Fig.~\ref{fig: dial_concept} (middle). More intuitively, the feature space can mix apples of the simulator with oranges of the target domain and yet have have a perfect alignment of marginal distributions.

To this end, we present a \textbf{domain-invariant representation learning} (DIRL) algorithm that resolves the ambiguity in domain adaptation by leveraging upon a
\begin{wrapfigure}{r}{8.8cm}
\vspace{-12pt}
\begin{center}
\includegraphics[trim={0.cm 1.0cm 0cm 0.cm},clip,scale=0.53]{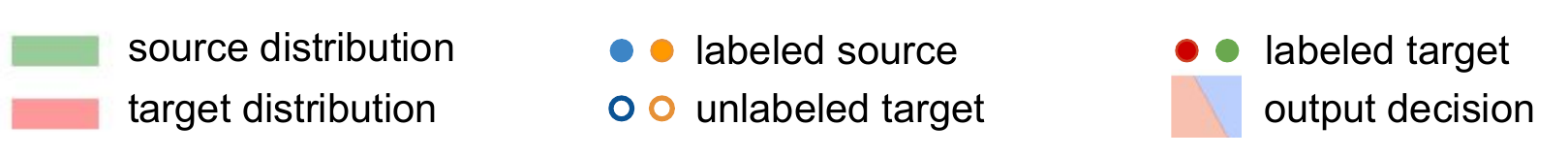}
\hrule
\includegraphics[trim={0.6cm 0.5cm 0.5cm 0.2cm},clip,scale=0.225]{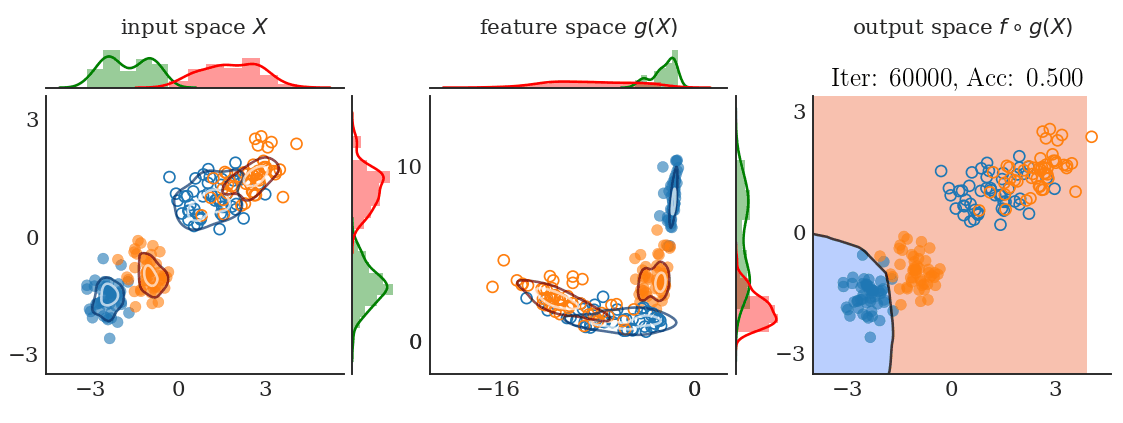}
\vspace{-0.32cm}
\captionsetup{labelformat=empty}
\caption{\scriptsize{conventional supervised learning on labeled source data}}
\hrule
\includegraphics[trim={0.7cm 0.5cm 0.5cm 0.0cm},clip,scale=0.225]{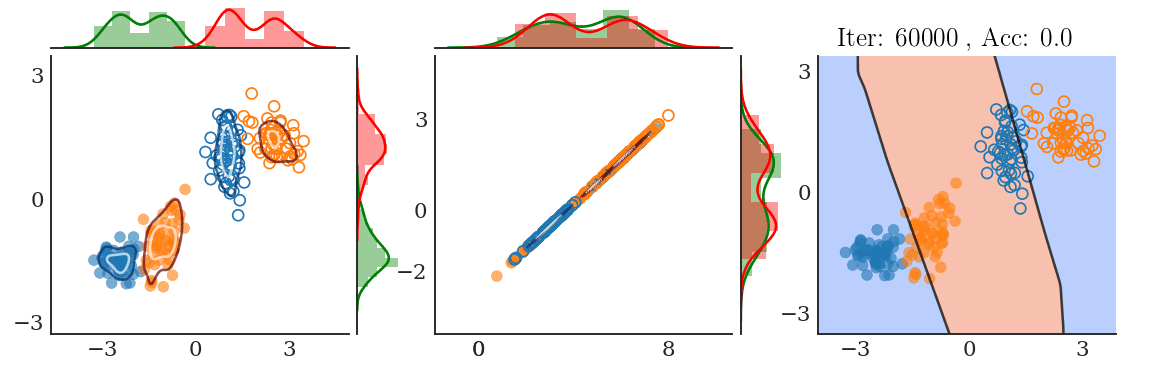}
\vspace{-0.8cm}
\caption{\tiny{\textit{\underline{\textbf{cross-label match}} with swapped labels across the decision boundary}}}
\includegraphics[trim={0.7cm 0.5cm 0.5cm 0.2cm},clip,scale=0.225]{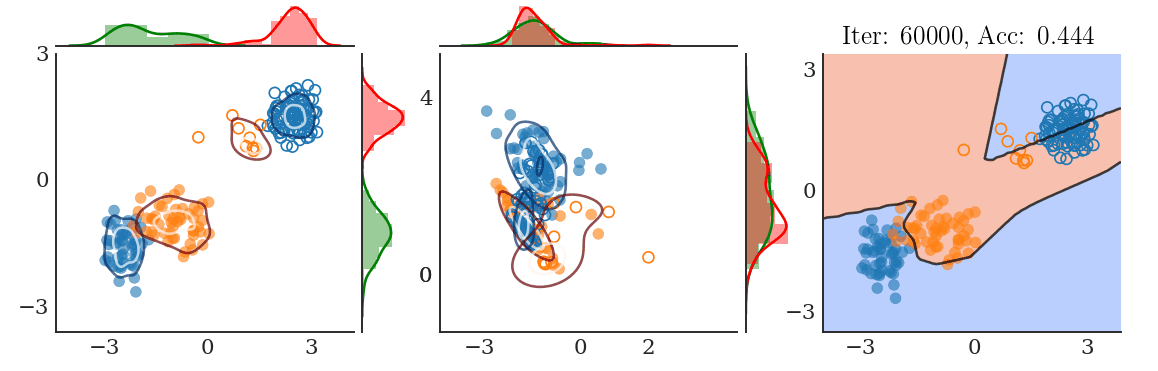}
\vspace{-0.7cm}
\caption{\tiny{\textit{\underline{\textbf{label-shift}} with uneven mixing of class boundaries}}}
\vspace{-0.25cm}
\caption{\scriptsize{conventional domain alignment with marginal distributions}}
\hrule
\includegraphics[trim={0.7cm 0.5cm 0.5cm 0.0cm},clip,scale=0.225]{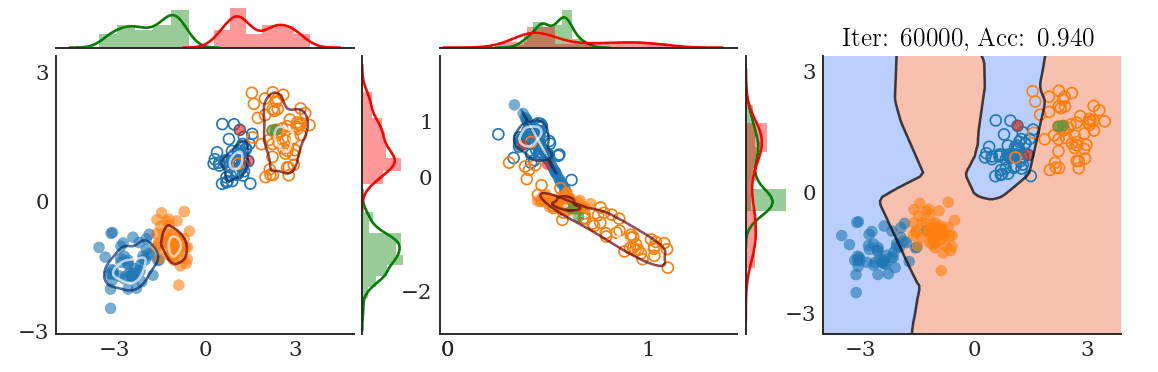} 
\vspace{-0.65cm}
\caption{\scriptsize{DIRL: domain and policy alignment with marginal and conditional distributions}}
\end{center}
\addtocounter{figure}{-5}
\vspace{-0.2cm}
\caption{\footnotesize{Domain-invariant representation learning on 2D synthetically generated data with $2$ classes (see App. for details): \textit{(top)} conventional supervised learning on the source domain does not generalize to the target domain drawn from a different distribution, \textit{(middle)} marginal distributions alignment only across domains can lead to negative transfer: a) cross-label match by swapping the class labels for the same target distribution, b) label-shift in decision boundary with class imbalance across domains, \textit{(bottom)} DIRL leverages upon a few labeled target examples to semantically align both the marginal and the conditional distributions. Note that the density plots at the top and the side indicate the marginal domain distributions, while the contour plots in the middle represent the class conditional distributions.}}\label{fig: dial_concept}
\vspace{-30pt}
\end{wrapfigure}
few labeled examples of the target domain to align both the marginal and the conditional distributions in the shared feature space with adversarial learning, and increases the inter-class variance and reduces the intra-class variance of the shared feature space with a triplet distribution loss. We apply the approach to vision-based robot decluttering, where a mobile robot picks objects from a cluttered floor and sorts them into respective bins~\citep{Gupta15,Gupta18,Tanwani_2019}. We seek to train the deep models with simulated images of object meshes and use a small number of labeled real world images to mitigate the domain shift between the image and the object distributions of the simulation and the real environments.

This paper makes three contributions: 1) we illustrate the limitations of cross label mismatch and/or label shift with unsupervised domain adaptation approaches that align marginal distributions in a shared feature space only, 2) we propose a domain-invariant representation learning (DIRL) algorithm to mitigate the covariate and conditional shift by aligning the marginal and the conditional distributions across domains, and making them disjoint with a novel triplet distribution loss in a shared feature space, and 3) we present experiments on MNIST benchmark domains and sim-to-real transfer of a single-shot object recognition model for mobile vision-based decluttering suggesting performance improvement over state-of-the-art domain adaptation methods.
\vspace{-10pt}
\section{Related Work}
\textbf{Domain Adaptation with Adversarial Learning:} A popular class of domain adaptation methods learn invariant transferable representations by matching feature and/or pixel distributions with similarity metrics like maximum mean discrepancy~\citep{Long_DA_16}, associative adaptation~\citep{Hausser_ADA_17}, reconstruction from latent features~\citep{Bousmalis_DSN_16,Ghifary_reconstructionDA_16}, and/or adversarial learning~\citep{Ganin_DANN_16,Bousmalis_PixelDA_16,Tzeng_adda_17,Hoffman_cycada_17,Zhu_cyclegan_17}. A common trend among these prior methods is to align the marginal distributions across the domains. This, however, does not guarantee that that the samples from the same class across the domains are mapped nearby in the feature space (see Fig.~\ref{fig: dial_concept} for an illustrative example). The lack of semantic alignment is a major limitation when the feature representation has conditional and/or label shift across the domains.

\textbf{Label and Conditional Domain Adaptation: }
An under explored line of work addresses matching the conditional or the label distributions in the feature space. Prior work in this direction make use of linear projections in the shared feature space~\citep{Long13,gong_icml_16}, estimate importance weights for label shift~\citep{Lipton_LS_18,Azizzadenesheli_19}, use domain discriminators with class specific adaptation~\citep{Hoffman_fcns_16,Chen_cdan_17}, combine class predictions with shared features as input to the discriminator~\citep{Long_cada_18}, maximize the conditional discrepancy with two classifiers~\citep{Saito_MCDDA_18}, augment domain adversarial training with conditional entropy regularization~\citep{shu_dirtt_18}, bound the density ratio with asymmetrical distribution alignment~\citep{Wu_19}, and align the weighted source and target distribution with importance weights under generalized label shift~\citep{Combes_2020}. 

Learning class conditional distributions of the unlabeled target domain requires pseudo-labels to encourage a low-density separation between classes~\citep{Kang_cvpr_19}. Tuning the thresholds for reliable prediction of pseudo-labels on unlabeled target data using only the source domain can be cumbersome and domain-specific. Semi-supervised methods provide a practical alternative when target data is scarce (for example, when acquired via kinesthetic or teleoperation interfaces). Prior examples of semi-supervised domain adaptation include coupling of domain and class input to the discriminator~\citep{Motiian_fada_17} and minimax conditional entropy optimization~\citep{Saito_ssda_19}.



In comparison to these approaches, the proposed domain-invariant represent learning approach provisions for both marginal and conditional alignment of domain distributions, while encouraging the feature representation of each class to be disjoint with a triplet distribution loss. We leverage upon a few target examples that stabilizes adaptation to challenging domains with large domain shift.

\textbf{Sim-to-Real Transfer: }Perception and control policies learned in simulation often do not generalize to the real robots due to the modeling inaccuracies. Domain randomization methods~\citep{Tobin_sim2real_17,Peng_sim2real_17,Chebotar_simopt_18,James_simlearn_18,Seita_2020} treat the discrepancy between the domains as variability in the simulation parameters, assuming the real world distribution as one such randomized instance of the simulation environment. In contrast, domain adaptation methods learn an invariant mapping function for matching distributions between the simulator and the robot environment. Related examples to the work presented here include using synthetic objects to learn a vision based grasping model~\citep{Saxena_grasping_08,mahler_dexnet2_17}, sim-to-real transfer of visuomotor policies for goal-directed reaching movement by adversarial learning~\citep{zhang_adda_grasp_19}, adapting dynamics in reinforcement learning~\citep{Eysenbach_2020}, and adapting object recognition model in new domains~\citep{Saenko_daor_10,Chen_darcnn_18,zhu_cvpr_19}. This paper focuses on a sample-efficient approach to learn invariant features across domains while reducing the cost of labeling real data. We show state-of-the-art performance on MNIST benchmarks and sim-to-real transfer of a single-shot object recognition model used in vision based surface decluttering with a mobile robot.


\vspace{-50pt}
\section{Problem Statement}
\vspace{-15pt}
We consider two environments: one belonging to the simulator or source domain $D_{S}$ comprising of the dataset $\{(\mb{x}_i^{S}, \mb{y}_i^{S} )\}_{i=1}^{N_S}$ and the other belonging to the real or target domain $D_{T}$ samples $\{(\mb{x}_i^{T}, \mb{y}_i^{T})\}_{i=1}^{N_T}$ with a few labeled samples $N_S \gg N_T$. The samples are drawn from $\mathcal{X} \times \mathcal{Y}$, where $\mathcal{X}$ is the input space and $\mathcal{Y}$ is the output space and the superscripts $S$ and $T$ indicate the draw from two different distributions of source and target random variables $(X^{S} \times Y^{S})$ and $(X^{T} \times Y^{T})$ respectively. Each output labeling function is represented as $\pi: \mathcal{X} \rightarrow \mathbb{R}^{|\mathcal{Y}|}$,  with $\pi_S$ and $\pi_T$ as the output policy of the source and the target domain. Let $\mathcal{Z}$ denote the intermediate representation space induced from $\mathcal{X}$ by a feature transformation $g: \mathcal{X} \rightarrow \mathcal{Z}$, which is mapped to the output conditional distribution by $f: \mathcal{Z} \rightarrow \mathcal{Y}$ under the transformation $X \xrightarrow{\; g \;} Z \xrightarrow{\; f \;} Y$. The composite function $f \circ g $ defines the policy $\pi$ of a domain, and the loss of a candidate policy $\pi \in \mathcal{H}$ represents the disagreement with respect to the source policy $\pi_{S}$ on domain $D_S$ as $\epsilon_S(\pi) := \epsilon_S(\pi,\pi_S) = \mathbb{E}_{\mb{x} \sim D_S}\; \left[\vert \pi(\mb{x}) -  \pi_S(\mb{x})\vert \right]$. For example, loss function for classification on target domain is represented by $\epsilon_T(\pi,\pi_T) = \mathbb{E}_{\mb{x} \sim D_T}\; [\mathbb{I} (\pi(\mb{x}) \neq \pi_T(\mb{x}))]$, while for regression as $\mathbb{E}_{\mb{x} \sim D_T}\; [\Vert \pi(\mb{x}) - \pi_T(\mb{x})) \Vert_2]$. 

The goal is to learn the policy $\pi \in \mathcal{H}$ using the labeled source examples and a few or no labeled target examples such that the error on the target domain is low. We seek to minimize the joint discrepancy across the domains in a shared feature space and map the output policy on the shared features to minimize the target error. Denoting the joint distribution of input domain and output labels across source and target domain as $\Pr(X^{S}, Y^{S}) = \Pr(Y^{S} | X^{S}) \cdot \Pr(X^{S})$, and $\Pr(X^{T}, Y^{T}) = \Pr(Y^{T} | X^{T}) \cdot \Pr(X^{T})$ respectively, the DIRL objective is to minimize,
\begin{align}
& \scalemath{0.83}{\left | \; \log \Pr(X^{S}, Y^{S}) - \log \Pr(X^{T}, Y^{T}) \;  \right | = 
\left | \; \log \Pr(Y^{S} \; | \; X^{S}) + \log \Pr(X^{S}) \; \right | -  \left | \; \log \Pr(Y^{T} \; | \; X^{T}) + \log \Pr(X^{T}) \; \right |,} \nonumber \\
 & = \left | \; \log \Pr(X^{S}) - \log \Pr(X^{T}) \; \right | \; + \; \left | \; \log \Pr(Y^{S} \; | \; X^{S}) - \log \Pr(Y^{T} \; | \; X^{T}) \; \right |, \nonumber \\
& \approx \left |\; g(X^{S}) - g(X^{T}) \; \right | \; + \; \left | \; f \circ g (X^{S}) - f \circ g (X^{T}) \; \right |, \nonumber \\
&= \underbrace{d_{\Pr\left(g(X)\right)}\left(D_S^{Z}, D_T^{Z}\right)}_{\text{marginal discrepancy}} \; \; +  \; \; \underbrace{d_{\Pr\left(Y|g(X)\right)}\left(D_S^{Y}, D_T^{Y}\right)}_{\text{conditional discrepancy}}.
\end{align}

The joint discrepancy consequently depends on both the \textbf{marginal discrepancy} and the \textbf{conditional discrepancy} between the source and the target domains defined in terms of a probability distance measure, and is minimized by aligning both the marginal and conditional distributions.




\begin{definition}
\label{def-domain-alignment}
\textit{\textbf{Marginal Distributions Alignment: } Given two domains $D_S=\{(\mb{x}_i^{S}, \mb{y}_i^{S} )\}_{i=1}^{N_S}$ and $D_T=\{(\mb{x}_i^{T}, \mb{y}_i^{T})\}_{i=1}^{N_T}$ drawn from two different distributions $\Pr(X^S, Y^{S}) \neq \Pr(X^T, Y^{T})$ with non-zero covariate shift $\textsc{KL}\left(\Pr(X^{S}) \Vert \Pr(X^{T})\right) > 0$, marginal alignment corresponds to finding the feature transformation $g: \mathcal{X} \rightarrow \mathcal{Z}$ such that the discrepancy between the transformed marginal distributions is minimized, i.e., $\Pr\left(g(X^{S})\right) = \Pr\left(g(X^{T})\right)$.}
\end{definition}
\begin{definition}
\label{def-class-alignment}
\textit{\textbf{Conditional Distributions Alignment: } Given two domains $D_S=\{(\mb{x}_i^{S}, \mb{y}_i^{S} )\}_{i=1}^{N_S}$ and $D_T=\{(\mb{x}_i^{T}, \mb{y}_i^{T})\}_{i=1}^{N_T}$ drawn from random variables $(X^{S} \times Y^{S})$ and $(X^{T} \times Y^{T})$ with different output conditional probability distributions $\Pr\left(Y^{S} \; |\;  X^{S}\right) \neq \Pr\left(Y^{T} \; | \; X^{T}\right)$, conditional alignment corresponds to finding the transformation $X \xrightarrow{\; g \;} Z \xrightarrow{\; f \;} Y$ such that the discrepancy between the transformed conditional distributions is minimized, i.e., $\Pr\left(Y^{S} \; | \; g(X^{S})\right) = \Pr\left(Y^{T} \; | \; g(X^{T})\right)$.}
\end{definition}
Note that the methods aligning marginal distributions only implicitly assume the same output conditional distribution $\Pr\left(Y \; | \; X\right)$ across the domains for the adaptation to be effective. This assumption of different marginal distribution across domains, but similar conditional distribution is known as \textit{covariate shift}~\citep{Candela_DS_09}. Several methods attempt to find invariant transformation to mitigate covariate shift such that $\Pr\left(g(X)\right)$ is similar across the domains by minimizing a domain discrepancy measure~\citep{Ganin_DANN_16,Tzeng_adda_17,Saito_MCDDA_18}. However, it is not clear if the assumption of $\Pr\left(Y \; | \; g(X)\right)$ also remains the same across the domains after the transformation. As the target labels may not be available in unsupervised domain adaptation, the class conditional distributions are naively assumed to be true under the transformation $\Pr\left(Y^{S} \; | \; g(X^{S})\right) = \Pr\left(Y^{T} \; | \; g(X^{T})\right)$. In this paper, we consider the joint marginal and conditional distributions alignment problem across the domains. Note that it is a non-trivial problem since we have access to only a few labels in the target domain. We consider the problem in a semi-supervised setting to resolve the ambiguity in conditional distributions alignment with a few labeled target examples.

\section{Theoretical Insights and Limitations}
Theoretical insights of a family of domain adaptation algorithms are based on the hypothesis that a provable low target error can be obtained by minimizing the marginal discrepancies between two classifiers~\citep{Ben-David_datheory_10,Zhao_DA_19,Johansson_DA_19,zhang_uda_2019,Li_2020}.
\begin{theorem}\textsc{(Ben-Davide et al.~\citep{Ben-David_datheory_10})}. Given two domains $D_S$ and $D_T$, the error of a hypothesis $\pi \in \mathcal{H}$ in the target domain $\epsilon_T(\pi)$ is bounded by the sum of: 1) the error of the hypothesis in the source domain, 2) the marginal discrepancy of the hypothesis class between the domains $d_{\mathcal{H}\Delta\mathcal{H}}(D_S^{Z}, D_T^{Z}) := 2 \sup_{\pi,\pi^{'} \in \mathcal{H}}| \epsilon_S(\pi,\pi^{'}) - \epsilon_T(\pi,\pi^{'}) |$, and 3) the best-in-class joint hypothesis error $\lambda_{\mathcal{H}} = \min_{\pi \in \mathcal{H}} \left[\epsilon_S(\pi) + \epsilon_T(\pi)\right]$,
\begin{equation}
\epsilon_T(\pi) \leq \epsilon_S(\pi) + \frac{1}{2} d_{\mathcal{H}\Delta\mathcal{H}}(D_S^{Z}, D_T^{Z}) + \lambda_{\mathcal{H}}.
\end{equation}
\end{theorem}
The $d_{\mathcal{H}\Delta\mathcal{H}}$ divergence can be empirically measured by training a classifier that discriminates between source and target instances, and subsequently minimized by aligning the marginal distributions between (unlabeled) source and target instances. The joint hypothesis error $\lambda_{\mathcal{H}}$ is widely assumed to be small, i.e., there exists a policy that performs well on the induced feature distributions of the source and the target examples after marginal alignment. More generally, the optimal joint error $\lambda_{\mathcal{H}}$ represents the cross-domain performance of the optimal policies $\min\{\epsilon_S(\pi_T), \epsilon_T(\pi_S)\}$. Hence, the upper bound on the target error can more appropriately be represented as,
\begin{equation}
\epsilon_T(\pi) \leq \epsilon_S(\pi) + d_{\mathcal{H}\Delta\mathcal{H}}(D_S^{Z}, D_T^{Z}) + \min\{ \mathbb{E}_{D_S} |\pi_S - \pi_T|,\mathbb{E}_{D_T} |\pi_S - \pi_T|  \}.\label{eq: da_bound}
\end{equation}

In Fig.~\ref{fig: dial_concept} (middle) showing conventional domain adaptation by aligning marginal distributions, the cross-overlapping area between class categories in the shared feature space (blue and orange) represents the optimal joint error given by $\epsilon_T(\pi_S, \pi_T)$ or $\epsilon_S(\pi_T, \pi_S)$. High joint error signifies that the conventional marginal alignment approaches fail in the presence of conditional or label shift between the domains (see also ~\citet{Zhao_DA_19} and ~\citet{zhang_uda_2019}). 
Consequently, we highlight two main limitations of aligning marginal distributions only with unsupervised domain adaptation as highlighted in Fig.~\ref{fig: dial_concept}: 1) \textbf{cross label matching}: labels of different classes are swapped in the shared feature space, and 2) \textbf{label shift}: the class distributions across the source and the target domains are imbalanced, leading to samples of one class being mixed with another class. In this work, we align both the marginal and the conditional discrepancies across domains and use a few labeled target samples to avoid ad-hoc mixing of class categories and negative transfer with cross label assignments.

\section{DIRL Algorithm}
\begin{figure}[!tbp]
\begin{center}
\includegraphics[trim={0.32cm 0.1cm 0.0cm 0.0cm},clip,scale=0.375]{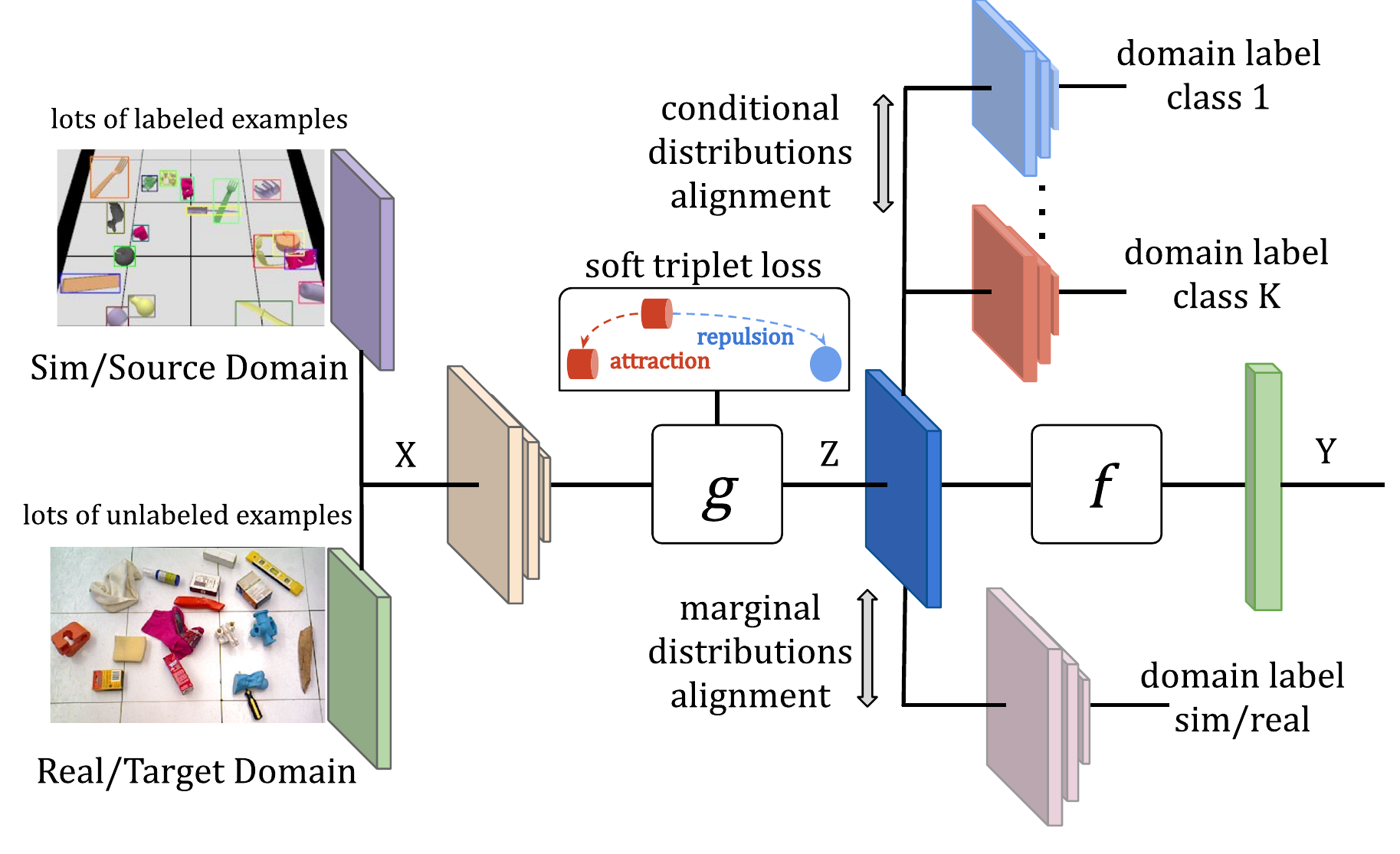} 
\vline
\includegraphics[trim={0cm -1cm 0.0cm 0cm},clip,scale = 0.27]{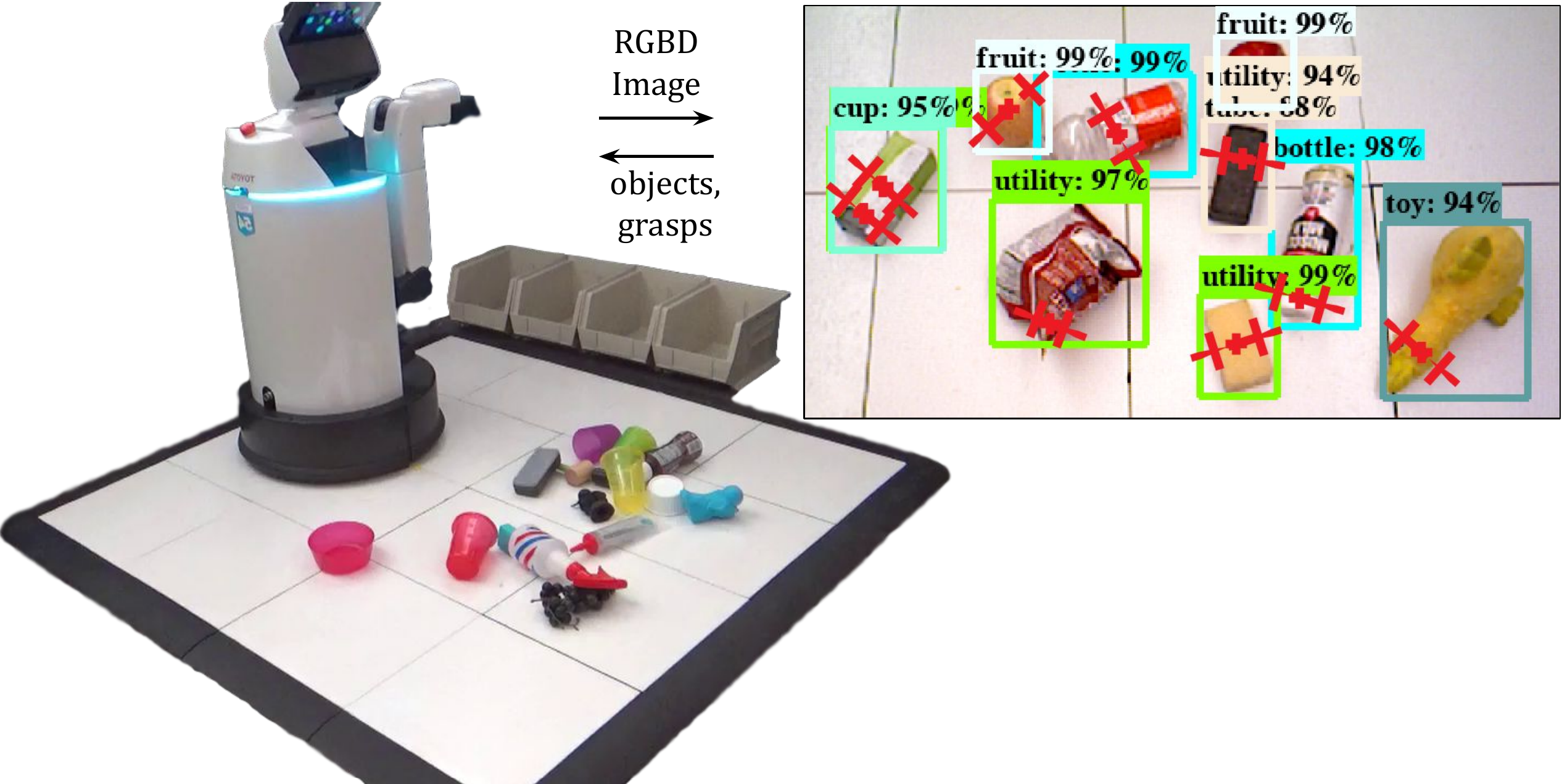}
\end{center}
\caption{\footnotesize{\textit{(left)} DIRL aligns marginal and conditional distributions of source and target domains, and uses a soft metric learning triplet loss to make the feature distributions disjoint in a shared feature space, \textit{(right)} experimental setup for decluttering objects into bins with HSR.}}\label{fig: dial_MA_CA}
\end{figure}

The central idea of the DIRL approach is to minimize the joint distribution error in a shared feature space by aligning the marginal and the conditional distributions across domains. To align the marginal distributions, we use a domain classifier to discriminate between the shared features of the source and the target domains. The shared features are adapted to deceive the domain classifier with adversarial learning~\citep{Ganin_DANN_16,Tzeng_adda_17,shu_dirtt_18}. Aligning the conditional distributions, however, is non-trivial due to the lack of labeled target examples. We align the conditional distributions in a semi-supervised manner by using class-wise domain discriminators for each class in a similar spirit to~\citet{Chen_cdan_17}. We further use a novel triplet distribution loss to make the conditional distributions disjoint in the feature space. The overall architecture of DIRL is shown in Fig.~\ref{fig: dial_MA_CA}.

\textbf{Marginal Distributions Alignment: }
Given the joint distribution $\Pr(X,Y) = \Pr(Y \vert X) \cdot \Pr(X)$, we align the marginal distributions of the transformed source and target domain with adversarial learning. The generator $g(X)$ encodes the data in a shared feature space, and the discriminator $D(X)$ predicts the binary domain label whether the data point is drawn from the source or the target distribution. The discriminator loss $\mathcal{L}_{ma}(g,D)$ is optimized using domain labels,
\begin{equation}
\min_D \; \mathcal{L}_{ma} \Bigl(g \left(\mb{x}_s, \mb{x}_t\right), D \left(\mb{x}_s, \mb{x}_t\right)\Bigr) =  -\mathbb{E}_{\mb{x}_s \sim X_s} \left[ \log D\left(g( \mb{x}_s)\right) \right] - \mathbb{E}_{\mb{x}_t \sim X_t} \left[ \log \left( 1 - D\left(g( \mb{x}_t)\right) \right) \right].
\end{equation}
The generator subsequently adopts the target domain features to confuse the discriminator with inverted domain labels to avoid vanishing gradients~\citep{Tzeng_adda_17}. Note that the gradient reversal layer can also be used~\citep{Ganin_DANN_16}. The generator loss $\mathcal{L}_{ma}(g, D)$ adapts the feature extractor,
\begin{equation}
\scalemath{1.0}{\min_g \; \mathcal{L}_{ma} \Bigl(g \left(\mb{x}_t\right), D\left(\mb{x}_s, \mb{x}_t\right)\Bigr)= -\mathbb{E}_{\mb{x}_t \sim X_t} \left[ \log D\left(g( \mb{x}_t)\right) \right]}.
\end{equation}
Without loss of generality, we denote the adaptation of the feature extractor with respect to the target data $\mb{x}_t$ only. The objective is optimized in a minimax fashion where the discriminator maximizes the empirical marginal discrepancy for a given features distribution, and the feature extractor minimizes the discrepancy by aligning the marginal distributions.
 




\textbf{Conditional Distributions Alignment: }
Conditional distributions alignment can overcome the issues of cross-label matching and the shift in labeling distributions with aligning marginal distributions alignment only. Effective alignment of conditional distributions depends upon two factors: 1) prediction of target pseudo-labels, and 2) balanced sampling of features per class across domains.

To this end, we leverage upon a few labeled target examples and train the output network with labeled source and target examples using the cross-entropy loss,
\begin{equation}
\scalemath{0.9}{\mathcal{L}_{ca\_sc} \Bigl(f \circ g\left(\mb{x}_s, \mb{y}_s, \mb{x}_t, \mb{y}_t \right)\Bigr) = \mathbb{E}_{\mb{x}_s, \mb{y}_s \sim (X_s, Y_s)} \left[- \mb{y}_s\log f\left( g(\mb{x}_s) \right)\right] + \\ \mathbb{E}_{\mb{x}_t, \mb{y}_t \sim (X_t, Y_t)} \left[- \mb{y}_t\log f\left( g(\mb{x}_t) \right)\right].}
\end{equation} We predict the pseudo-labels for the unlabeled target data, $\mb{\hat{y}}_t = \argmax f\left( g(\mb{x}_t)\right)$, by querying the network pre-trained on labeled data at an earlier stage during training and retain only top-$n$ pseudo-labels of each class category based on their confidence. We sample with replacement to create a balanced mini-batch with half source and half labeled target examples, and augment the mini-batch with pool of pseudo-labeled target examples after the pre-training stage only. 

A minimax game with adversarial learning aligns the conditional distribution with a domain discriminator for each class. First, the class discriminator $C(X)$ estimates the conditional discrepancy with respect to the source and the target data for a fixed feature extractor. Second, the generator adapts the feature extractor to minimize the conditional discrepancy for a fixed discriminator. Formally, the adversarial loss for each (ground-truth and predicted) class $k = 1 \ldots \vert\mathcal{Y} \vert$ is,
\begin{align}
\scalemath{0.86}{\min_C \; \mathcal{L}_{ca_k}\left(g (\mb{x}_s^{(k)}, \mb{x}_t^{(k)}),C(\mb{x}_s^{(k)}, \mb{x}_t^{(k)}\right)} & \scalemath{0.86}{=  -\mathbb{E}_{\mb{x}_s^{(k)} \sim X_s} \left[ \log C\left(g( \mb{x}_s^{(k)})\right) \right] - \mathbb{E}_{\mb{x}_t^{(k)} \sim X_t} \left[ \log \left( 1 - C\left(g( \mb{x}_t^{(k)})\right) \right) \right],} \nonumber \\ 
\scalemath{0.86}{\min_g  \; \mathcal{L}_{ca_k} \left(g (\mb{x}_s^{(k)}, \mb{x}_t^{(k)}),C(\mb{x}_s^{(k)}, \mb{x}_t^{(k)}\right)} & \scalemath{0.86}{= -\mathbb{E}_{\mb{x}_t^{(k)} \sim X_t} \left[ \log C\left(g( \mb{x}_t^{(k)})\right) \right].}
\end{align} Minimizing  the conditional discrepancy penalizes the feature extractor to separate the cross domain overlap for each class to give low joint error $\epsilon_T(\pi_s, \pi_T)$ in Eq.~\eqref{eq: da_bound} for provably effective adaptation to the target domain.







\textbf{Triplet Distribution Loss: }
To increase the inter-class variance and reduce the intra-class variance, we introduce a variant of triplet loss~\citep{Schroff15,Rippel15} that operates on distributions of anchor, positive and negative examples, instead of tuples of individual examples (see also~\citep{Florence_thesis_2019}). Given $M$ labeled examples in a mini-batch, the loss posits that the Kullback-Leibler (KL)-divergence between anchor and positives distribution in the shared feature space is less than the KL-divergence between anchor and negatives distribution by some constant margin $\alpha_{\mathrm{tl}} \in \mathbb{R}_{+}$. Mathematically, 
\begin{equation}
\scalemath{0.76}{\mathcal{L}_{tl} = \overbrace{\sum_{a=1}^{M}}^{\text{all anchors}} \Biggl[  \overbrace{\frac{1}{M_p-1}\sum_{\substack{p=1 \\ p \neq a}}^{M_p}  \text{KL}\left(\mathcal{N} \left(\bar{g}(\mb{x}_a), \sigma^2 \right) \Big\Vert \; \mathcal{N} \left(\bar{g}(\mb{x}_p), \sigma^2 \right) \right)}^{\text{all positives}} \; - \; \overbrace{\frac{1}{M_n}\sum_{n=1}^{M_n} \text{KL}\left(\mathcal{N} \left(\bar{g}(\mb{x}_a), \sigma^2 \right) \Big \Vert \; \mathcal{N} \left(\bar{g}(\mb{x}_n), \sigma^2 \right) \right)}^{\text{all negatives}} \;  + \;  \alpha_{\mathrm{tl}}\Biggr]_{+}},
\end{equation} where $\{.\}_+$ is the hinge loss, $\bar{g}(\mb{x})$ is normalized to extract scale-invariant features similar to~\citep{Schroff15}, $\mathcal{N} \left(\bar{g}(\mb{x}_a), \sigma^2 \right)$ is short for distribution of examples,  $\left \{\mathcal{N} \left(\bar{g}(\mb{x}_i); \bar{g}(\mb{x}_a), \sigma^2 \right) = \frac{\exp^{-\frac{1}{\sigma^2} \Vert \bar{g}(\mb{x}_i) - \bar{g}(\mb{x}_a)\Vert_2^{2} }}{\sum_{j=1}^{K}\exp ^{-\frac{1}{\sigma^2} \Vert \bar{g}(\mb{x}_j) - \bar{g}(\mb{x}_a)\Vert_2^{2}}} \right\}_{i=1}^{K}$, with a Gaussian situated on normalized anchor example in the feature space $\bar{g}(\mb{x}_a)$, $M_p$ and $M_n$ are the number of positive and negative examples in a mini-batch, and $\sigma^2 \in \mathbb{R}$ is the hyper-parameter to control the variance of the Gaussian distribution. For each anchor, the positive examples belong to the same class as that of anchor, while the negative examples are sampled from other classes. The soft variant of triplet loss encourages the features distribution to be robust to outliers, while increasing the inter-class variance and reducing the intra-class variance across similar examples in the shared feature space.

\textbf{Overall Algorithm: } The overall DIRL algorithm comprises of the classification loss on the labeled source and target examples, marginal and conditional distributions alignment loss on the features generator, and triplet distribution loss on the labeled source and target examples. Given the weight coefficients $\lambda_i > 0$ of the respective losses, the overall loss function $\mathcal{L}_{\mathrm{DIRL}}$ that DIRL optimizes is, 
\begin{multline}
\scalemath{0.85}{\mathcal{L}_{\mathrm{DIRL}} = \lambda_1 \mathcal{L}_{ca\_sc} \Bigl(f \circ g \left(\mb{x}_s, \mb{y}_s, \mb{x}_t, \mb{y}_t \right) \Bigr) \; + \;  \lambda_2 \mathcal{L}_{ma}\Bigl(g \left(\mb{x}_t\right), D \left(\mb{x}_s, \mb{x}_t\right) \Bigr) \; + \;} \\  \scalemath{0.85}{\lambda_3  \sum_{k=1}^{\vert\mathcal{Y} \vert}\mathcal{L}_{ca_k}\Bigl(g (\mb{x}_t^{(k)}), D (\mb{x}_s^{(k)}, \mb{x}_t^{(k)}) \Bigr) + \lambda_4 \mathcal{L}_{tl}\Bigl(g\left(\mb{x}_s, \mb{y}_s, \mb{x}_t, \mb{y}_t \right) \Bigr).}
\end{multline} 

\section{Experiments, Results and Discussions}

In this section, we first benchmark the DIRL algorithm on digits domains, followed by sim-to-real transfer of vision-based decluttering with a mobile robot (see Appendix for 2D synthetic example in Fig.~\ref{fig: dial_concept}). We empirically investigate what representations transfer better with unsupervised approaches, and the effect of a few labeled target examples in transfer learning across domains. 


\begin{wraptable}{r}{10.5cm}
\vspace{-12pt}
\scriptsize
\caption{\footnotesize Average test accuracy on target domains of Digits datasets with unsupervised and semi-supervised domain adaptation. DIRL consistently performs well across all target domains in comparison to other baselines. *results from~\citep{Motiian_fada_17}.} \centering \label{tab: digits_dial}
\begin{tabular}{|c c ||c|c|c|c|c|c|}
\hline
 \multicolumn{2}{|c||}{\textbf{Methods}}  & \textbf{MNIST}$\rightarrow$ & \textbf{MNIST}$\rightarrow$ & \textbf{SVHN}$\rightarrow$ & \textbf{USPS}$\rightarrow$ & \textbf{USPS}$\rightarrow$ & \textbf{MNIST}$\rightarrow$\\ 
 &  & \textbf{MNISTM} & \textbf{USPS} & \textbf{MNIST} & \textbf{SVHN} & \textbf{MNIST} & \textbf{SVHN} \\ \hline \hline
\multicolumn{8}{|c|}{\textbf{unsupervised}}\\\hline
\multicolumn{2}{|c||}{\textbf{RDA}} & $0.534$ & $0.823$ & $0.627$ & $0.155$ & $0.715$ & $0.195$ \\ \hline
\multicolumn{2}{|c||}{\textbf{DANN}} & $0.798$ & $0.873$ & $0.674$ & $0.153$ & $0.751$ & $0.194$ \\ \hline
\multicolumn{2}{|c||}{\textbf{Triplet}} & $0.704$ & $0.886$ & $0.805$ & $0.180$ & $0.832$ & $0.212$ \\ \hline
\multicolumn{2}{|c||}{\textbf{ADA}} & $\textbf{0.895}$ & $0.212$ & $0.960$ & $0.256$ & $0.570$ & $\textbf{0.359}$ \\ \hline
\multicolumn{2}{|c||}{\textbf{MCD}} & $0.770$ & $\textbf{0.941}$ & $\textbf{0.978}$ & $\textbf{0.288}$ & $\textbf{0.932}$ & $0.330$ \\ \hline \hline 
\multicolumn{8}{|c|}{\textbf{semi-supervised}}\\\hline
\multirow{3}{*}{\textbf{DANN}} & \multicolumn{1}{|c||}{$1$} & $0.758$ & $0.872$ & $0.764$ & $0.151$ & $0.777$ & $0.228$ \\ 
& \multicolumn{1}{|c||}{$5$} & $0.763$ & $0.920$ & $0.793$ & $0.252$ & $0.850$ & $0.353$ \\
& \multicolumn{1}{|c||}{$10$} & $0.796$ & $0.923$ &$ 0.806$ & $0.354$ & $0.914$ & $0.405$ \\ \hline
\multirow{3}{*}{\textbf{FADA*}} & \multicolumn{1}{|c||}{$1$} & & $0.891$ & $0.728$ & $0.275$ & $0.811$ & $0.377$ \\ 
&\multicolumn{1}{|c||}{$5$} & $-$ & $0.934$ & $0.861$ & $0.379$ &  $0.911$ & $0.461$ \\
&\multicolumn{1}{|c||}{$7$} &  & $0.944$ & $0.872$ & $0.429$ & $0.915$ & $0.470$ \\  \hline
\multirow{3}{*}{\textbf{DIRL}} & \multicolumn{1}{|c||}{$1$} & $0.786$ & $0.894$ & $0.773$ & $0.400$ & $0.949$ & $0.495$ \\ 
& \multicolumn{1}{|c||}{$5$} & $0.941$ & $0.946$ & $0.864$ & $0.610$ & $0.945$ & $0.683$ \\
& \multicolumn{1}{|c||}{$10$} & $\textbf{0.948}$ & $\textbf{0.951}$ & $\textbf{0.903}$ & $\textbf{0.802}$ & $\textbf{0.962}$ & $\textbf{0.837}$ \\ \hline
\end{tabular}
\vspace{-8pt}
\end{wraptable}

\subsection{Digits Datasets }
We compare the DIRL approach with state-of-the-art methods including DANN~\citep{Ganin_DANN_16}, associative domain adaptation (ADA)~\citep{Hausser_ADA_17}, reconstruction based domain adaptation (RDA)~\citep{Ghifary_reconstructionDA_16}, MCD~\citep{Saito_MCDDA_18}, and FADA~\citep{Motiian_fada_17} on six source$\rightarrow$target benchmarks, namely: MNIST$\rightarrow$MNISTM, MNIST$\rightarrow$USPS, SVHN$\rightarrow$MNIST, USPS$\rightarrow$SVHN, USPS$\rightarrow$MNIST, and  MNIST$\rightarrow$SVHN. 

Results in Table~\ref{tab: digits_dial} suggest that the unsupervised methods aligning marginal distributions only often do not perform well when the target domain discrepancy increases relative to the source domain. MCD performs better among unsupervised approaches by minimizing the conditional discrepancy loss using two classifiers, but gives unsatisfactory results with challenging adaptation situations such as USPS$\rightarrow$SVHN and MNIST$\rightarrow$SVHN. DIRL addresses the large domain shift by aligning both the marginal and the conditional distributions using only a few target examples, and consistently outperforms the compared unsupervised and semi-supervised DANN and FADA approaches. As an example, MNIST $\rightarrow$ SVHN accuracy increases by $28.2\%$ from $1$-shot to $5$-shot and by $12.0\%$ from $5$-shot to $10$-shot target examples of each class. 

\subsection{Vision-Based Decluttering by Sim-to-Real Transfer }
Robots picking diversely shaped and sized novel objects in cluttered environments has a wide range of near-term applications in homes, schools, warehouses, offices and retail stores. We consider this scenario with a mobile Toyota Human Support Robot (HSR) that observes the state of the floor as a RGB image $\mb{I}_{t}^{c} \in \mathbb{R}^{640 \times 480 \times 3}$ and a depth image $\mb{I}_{t}^{d} \in \mathbb{R}^{640 \times 480}$. The robot recognizes the objects $\{o_{i}\}_{i=1}^{N}$ as belonging to the object categories $o_{i} \in \{1 \ldots |\mathcal{O}|\}$, and subsequently plans a grasp action $\mb{y}_t \in \mathbb{R}^{4}$ corresponding to the $3$D object position and the planar orientation of the most likely recognized object. After grasping an object, the robot places the object into appropriate bins (see Fig.~\ref{fig: dial_MA_CA} \textit{(right)} for an overview). 

In this work, we investigate the feasibility of vision-based decluttering with a mobile robot by sim-to-real transfer with the proposed DIRL approach in a semi-supervised manner. We simulate the decluttering environment in Pybullet similar to the setup in Fig.~\ref{fig: dial_MA_CA}, and collect $20,000$ synthetic RGBD images of cluttered object meshes on the floor, each containing $5-25$ objects drawn from a distribution of screwdriver, wrench, fruit, cup, bottle, assembly part, hammer, scissors, tape, toy, tube and utility object meshes collected from publicly available repositories. We vary the camera viewpoint, the background texture and color of object meshes in each image and store the ground-truth bounding box locations, object categories and analytically evaluated grasps for uniformly sampled antipodal pairs on the object meshes in an image. Additionally, we collect $212$ RGBD images with the HSR on $1.2$ sq. meter white tiled floor in a similar manner from a distribution of $102$ household and machine shop objects, and hand-label the bounding boxes and object categories.

\begin{wraptable}{r}{6.5cm}
\vspace{-12pt}
\scriptsize
\caption{\footnotesize Performance evaluation of domain-invariant object recognition by sim-to-real transfer on target test set using mean Average Precision (mAP), classification accuracy on synthetic test images $\mathrm{sim\_eval}$, real test images $\mathrm{real\_eval}$ and silhouette score $\mathrm{SS}$. DIRL performs better than other compared approaches across both domains.} \centering \label{tab: dior_dial}
\begin{tabular}{|c||c|c|c|c|}
\hline
Methods & \textbf{mAP} & $\mathrm{\textbf{sim\_eval}}$ & $\mathrm{\textbf{real\_eval}}$ & $\mathrm{\textbf{SS}}$\\ \hline \hline
\textbf{Sim Only} & $0.13$ & $95.7$ & $26.8$  & $0.08$\\ \hline 
\textbf{Real Only} & $0.62$ & $24.6$ & $85.9$& $0.42$ \\ \hline 
\textbf{Sim $+$ Real} & $0.33$& $95.5$ & $70.6$ & $0.19$ \\ \hline
\textbf{Triplet} & $0.52$& $94.5$& $76.2$ & $0.35$ \\ \hline
\textbf{DANN} & $0.61$& $93.2$& $84.4$ & $0.30$ \\ \hline
\textbf{MCD} & $0.65$& $94.1$& $89.6$ & $0.48$ \\ \hline \hline
\textbf{DIRL} & $\mb{0.69}$& $\textbf{94.2}$& $\textbf{91.0}$ & $\textbf{0.69}$ \\ \hline
\end{tabular}
\vspace{-8pt}
\end{wraptable}

We modify the single shot multi-box detector (SSD)~\citep{Lin17} with focal loss and feature pyramid as the base model for domain-invariant object recognition. Domain classifiers for marginal and conditional discrepancy are added on top of feature pyramid networks (see supplementary materials for architecture and training details). Results with $127$ labeled target examples on test set are summarized in Table~\ref{tab: dior_dial}. Note that the Silhouette score (SS) metric in Table~\ref{tab: dior_dial} measures the tightness of a cluster relative to the other clusters without using any labels (unsupervised); while the classification accuracy and the mAP are supervised metrics that use the target test set labels. We observe that the object recognition model trained on synthetic data only gives poor performance on real data with $26.8\%$ accuracy, in comparison to $85.9\%$ accuracy obtained with training on real labeled data only. Naively combining the synthetic and real data in a mini-batch is also sub-optimal. Using triplet loss on labeled source examples preserves the structure of the features for transfer to real examples. DANN improves performance in both domains by aligning marginal distributions. MCD further improves the performance in domain adaptation with conditional alignment of distributions. DIRL outperforms the compared approaches by combining marginal and conditional distributions alignment with triplet distribution loss.

\textbf{Decluttering with Toyota HSR: }We test the performance on the Toyota HSR for picking objects from the floor and depositing them in target bins as shown in Fig.~\ref{fig: dial_MA_CA} \textit{(right)}. We load $5 - 25$ objects in a bin from a set of $65$ physical objects and drop them on the floor in front of the robot. The objects may overlap after dropping; a pushing primitive is used to singulate the cluttered objects if the overlap is more than a threshold. The domain-invariant object recognition model gives $89.4 \%$ accuracy in real experiments. The cropped depth image from the output bounding box of the object recognition model is fed as input to the grasp planning model trained on simulated depth images, adapted from~\citep{Tanwani_2019,staub_hsr_2019,mahler_dexnet2_17}. The grasping network gives $86.5 \%$ accuracy of picking up the identified object. Without using the grasping network and only grasping orthogonal to the principal axis of the point cloud of the predicted object location gives $76.2 \%$ accuracy. We observe that the robot performs well in grasping compliant objects and objects with well-defined geometry such as cylinders, screwdrivers, tape, cups, bottles and utilities; while assembly parts and small bowls in inverted pose induced repeated failures in grasping the target object (see supplementary materials for details).

\section{Conclusion} 
\label{sec:conclusion}
In this paper, we have presented a sample-efficient domain-invariant representation learning approach for adapting deep models to a new environment. The proposed DIRL approach overcomes the ambiguity in transfer with unsupervised approaches by mitigating both the marginal and the conditional discrepancies across domains with a small amount of labeled target examples, and uses a triplet distribution loss to make the feature distributions disjoint in a shared feature space. Experiments with digit domains yield state-of-the-art performance across challenging transfer scenarios with large domain shift, while vision-based decluttering with a mobile robot suggest the feasibility of grasping diversely shaped and sized novel objects by sim-to-real transfer. In future work, we plan to investigate domain-invariance while deploying models across multiple environments with a fleet of robots~\cite{tanwani_rilaas_20} and close the real-to-sim loop in transferring models across new environments.




\clearpage
\acknowledgments{The research work was performed at UC Berkeley, in collaboration with the AutoLab, the Berkeley Artificial Intelligence Research (BAIR), and the Real-Time Intelligent Secure Execution (RISE) Lab. Daniel Zeng and Matthew Trepte helped with the design and the experiments of the digits and the object recognition domains respectively. Thanks to Kate Sanders, Yi Liu, Daniel Seita, Lerrel Pinto, Trevor Darrell, and Ken Goldberg for their feedback and suggestions.
}


\small
\bibliography{bibliography_cvpr}  
\newpage
\appendix
\section{Experimental Details}
\subsection{2D Synthetic Example}

We consider a $2$-dimensional problem comprising of $2$ classes. Source data is generated from Gaussian distributions with means centered around $[-2.5, -1.5]$ and $[-1.0, -1.0]$ respectively for the two classes. Covariance matrices for all the Gaussians are represented by: $\sigma^2 \mb{I} + \mb{W}\mb{W}^{\trsp}$ with $\sigma^2=0.1$ and $\mb{W} = \; $0.25$ \; \mathcal{N}(\mb{0}, \mb{I})$. Similarly, Target data is drawn from Gaussian distributions with respective class means centered around $[1.0, 1.0]$ and $[2.5, 1.5]$. We sample $1000$ instances from the source and the target distribution each for training, and $100$ instances from the target distribution are used for the testing the learned model. We use the same weight for all the constituent loss functions with $\lambda_1 \ldots \lambda_4 = 1.0$. Network architecture comprises of $3$ hidden layers of $7$ neurons with ReLU activation for each of the shared feature space, output classifier, domain discriminator and class-conditional domain discriminator. During training, we use a mini-batch of $80$ samples comprising of half source, labeled target (if applicable) and unlabeled target examples. For conditional domain discriminators, we sub-sample a new mini-batch comprising of half labeled source and half labeled target examples for each class. We use Adam optimizer with a learning rate of $0.0001$ for $60$K iterations. Note that we do not use labeled target instances for classification loss, and only use them for triplet distribution and conditional discriminator loss, in order to better analyze the effect of constituent losses on the target accuracy. Results are summarized in Fig.~\ref{fig: dial_concept} and the animations are available on the project website: \ajay{\url{https://sites.google.com/view/dirl}}


\subsection{Digits Datasets }
\begin{figure}[!tbp]
\begin{center}
\includegraphics[trim={0.cm 0.25cm 0cm 0.cm},clip,scale=0.21]{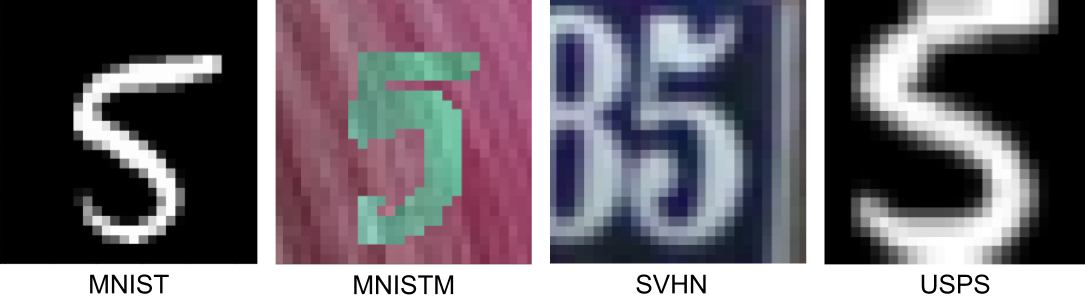}
\includegraphics[trim={0cm 0cm 0.0cm 0cm},clip,scale = 0.5]{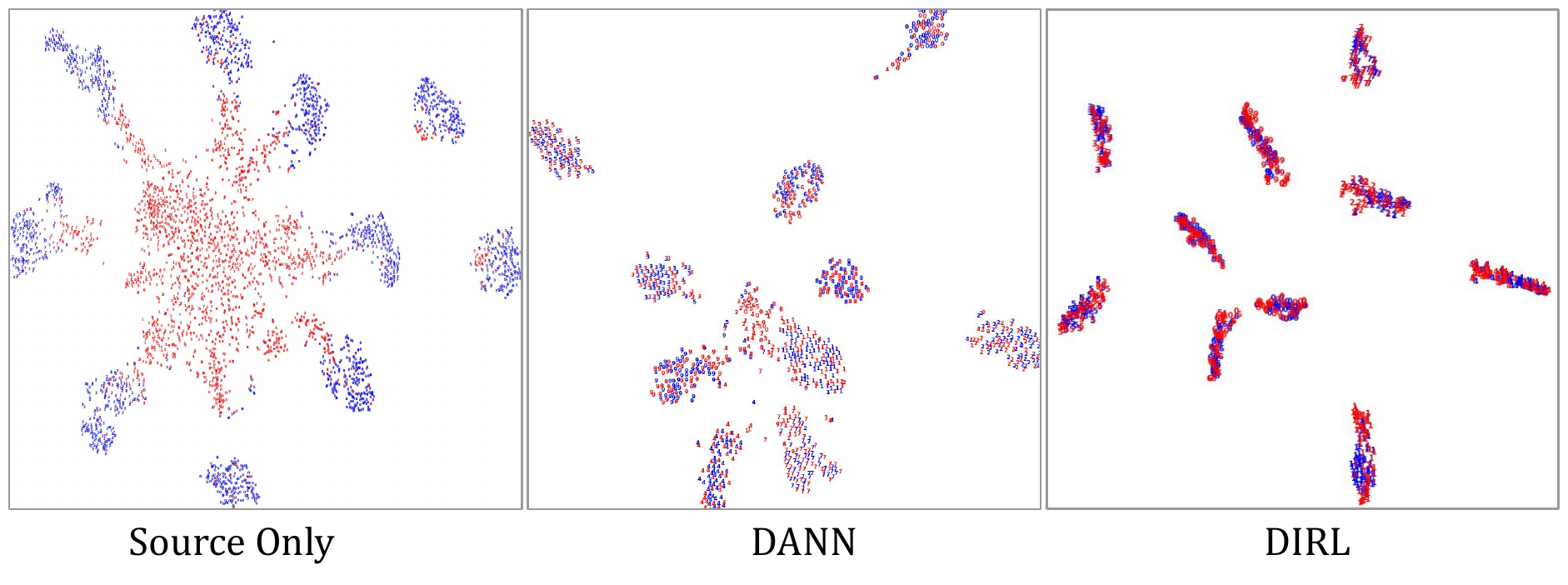}
\end{center}
\caption{\small{\textit{(top)} Sample image of label $5$ from MNIST, MNISTM, SVHN and USPS digits datasets, \textit{(bottom)} T-SNE visualization of \textbf{MNIST}$\rightarrow$\textbf{MNISTM} (source in blue, target in red). DIRL compactly clusters the class distributions across datasets for transfer learning in comparison to DANN and source only transfer.}}\label{fig: dirl_digits}
\end{figure}

We choose four commonly used digits datasets for benchmarks (see Fig.~\ref{fig: dirl_digits}): MNIST~\cite{lecun_mnist_2010}, MNISTM~\cite{Ganin_DANN_16}, USPS~\cite{hull_usps_94}, and SVHN~\cite{netzer_svhn_11}. We select first dataset as the source and the second one as the target dataset for adaptation in the following configurations: \textbf{MNIST}$\rightarrow$\textbf{MNISTM}, \textbf{MNIST}$\rightarrow$\textbf{USPS}, \textbf{SVHN}$\rightarrow$\textbf{MNIST},
\textbf{USPS}$\rightarrow$\textbf{SVHN},
\textbf{USPS}$\rightarrow$\textbf{MNIST},
\textbf{MNIST}$\rightarrow$\textbf{SVHN}.

\begin{wraptable}{r}{6.5cm}
\vspace{-12pt}
\scriptsize
\caption{\footnotesize Digits Dataset Instances.} \centering \label{tab: dior_dial}
\begin{tabular}{|c||c|c|}
\hline
Dataset & \textbf{Train Instances} & \textbf{Test Instances} \\ \hline \hline
\textbf{MNIST} & $55,000$ & $10, 000$ \\ \hline 
\textbf{MNISTM} & $55, 000$ & $10, 000$ \\ \hline 
\textbf{SVHN} & $73, 257$& $26, 032$ \\ \hline
\textbf{USPS} & $7, 291$& $2, 007$ \\ \hline
\end{tabular}
\vspace{-8pt}
\end{wraptable}

Experimental details are as follows: Input dimension of all datasets are reshaped to $28 \times 28 \times 3$, output classifier dimension is $10$ corresponding to the unit digits. Network architecture consists of the shared feature space: $3$ sets of convolution layers with $32, 64, 128$ hidden layers each separated by a max pool layer with stride of length $2$. The output of the final layer is flattened to $256$ dimensional space and fed to the output classification network, domain discriminator and conditional domain discriminator with $3$ dense layers of size $100, 50$ and the output dimensions of $10, 2, 2$ respectively.

We use the same weight for all the constituent loss functions with $\lambda_1 \ldots \lambda_4 = 1.0$. We perform experiments with $\{1, 5, 10\}$ labeled instances to evaluate the proposed approach. Overall batch size comprises of half-source and half-target examples. Source examples are always labeled, while target examples are 
$25\%$ labeled and $75\%$ unlabeled within a mini-batch. No pseudo-labels are used for the target examples with the digits experiments. Adam optimizer with learning rate of $0.001$ for $10, 000$ iterations to minimize the constituent loss functions. We also attempted instance normalization with this setup similar to~\cite{shu_dirtt_18}, but it did not have any significant effect on the results.

We compare the average test accuracy of the target domains in unsupervised and semi-supervised setting. Methods aligning marginal distributions only such as DANN often do not perform well due to lack of conditional alignment across source domains. Metric learning with triplet loss on the source domain increases the separation between class categories, which helps in transfer to the new environment. Adding reconstruction loss on top of DANN to force the feature transformation to be invertible decreases the performance on the target domain. This performance degradation is likely due to the additional constraints of having distinct features for each sample, making it difficult to align the marginal distributions as also suggested in~\citet{Johansson_DA_19}. Associative domain adaptation imposes a cyclic loss to bring source and target examples close in the shared feature space, however, yields unreliable performance across datasets. MCD performs better across unsupervised baselines by minimizing the conditional discrepancy loss using two classifiers, however, gives unsatisfactory results with challenging adaptation situations such as \textbf{USPS}$\rightarrow$\textbf{SVHN} and \textbf{MNIST}$\rightarrow$\textbf{SVHN} like other unsupervised domain adaptation methods.

DIRL addresses the limitations of the existing approaches and outperforms other unsupervised and semi-supervised approaches with DANN, FADA in $1$-shot, $5$-shot and $10$-shot scenario. DIRL uses a few target labels for effective conditional distributions alignment (see Fig.~\ref{fig: dirl_digits} for a qualitative comparison). Results presented in this paper do not make use of the pseudo-labeled target data. Note that DIAL performs well across all datasets, especially on challenging problems of \textbf{USPS}$\rightarrow$\textbf{SVHN} and \textbf{MNIST}$\rightarrow$\textbf{SVHN} with large domain shift by aligning the marginal and the conditional distributions using only a few samples for the target class. Specifically, \textbf{MNIST $\rightarrow$ SVHN} increases $28.2\%$ from $1$-shot to $5$-shot and $12.0\%$ from $5$-shot to $10$-shots.

\subsection{Vision-Based Decluttering by Sim-to-Real Transfer }
\textbf{Simulation and Real Dataset: }We simulate the decluttering environment in a Pybullet simulator. The simulated dataset comprises of $20,000$ synthetic RGB and depth images of cluttered object meshes on floor. Each image randomly contains between $5-25$ objects that are split across $12$ categories, namely screwdriver, wrench, fruit, cup, bottle, assembly part, hammer, scissors, tape, toy, tube and utility. The object meshes are collected from Turbosquid, Kit, 3dNet, ShapeNet repositories. We use domain randomization to vary the camera viewpoint, background texture and color of object meshes in each image, and store the ground-truth bounding box locations, object categories, segmentation masks and analytically evaluated grasps for uniformly sampled antipodal pairs on the object meshes in an image. 

The physical dataset comprises of $212$ real RGB and depth images collected with the Toyota HSR looking at $1.2$ sq. meter white tiled floor from a distribution of $102$ household and machine shop objects, and hand-label the bounding boxes and object categories. We hand-label the bounding boxes and object categories similar to the synthetic classes above.

\textbf{Object Recognition Network: }We use the MobileNet-Single Shot MultiBox Detector (SSD) \cite{Liu15,Lin17} algorithm with focal loss and feature pyramids as the base model for object recognition. The input RGB image $\in \mathbb{R}^{640 \times 480 \times 3}$ is fed to a pre-trained VGG$16$ network, followed by feature resolution maps and a feature pyramid network. The feature pyramid network produces feature maps across $5$ resolutions: $80 \times 80$, $40 \times 40$, $20 \times 20$, $10 \times 10$, $5 \times5$ that are concatenated before being fed to the class prediction and box prediction networks. 

We modify the base model by flattening the output of the class predictions with background features and adding domain and class discriminators on top. Domain discriminator consists of three fully connected layers of size $1024, 200, 100$ and $2$ output neurons for classifying simulator vs real images. Class conditional domain discriminator consists of two fully connected layers of size $200, 100$ before the output layer of $2$ neurons. We use an input batch size of $8$ with $4$ clones that are split across half-sim, $25\%$ labeled and $25\%$ unlabeled real data. We sample $128$ most like class predictions (without background) in the same proportion as the batch size for triplet distribution loss, marginal and conditional discriminators. After pretraining the network for $4000$ iterations, we assigned pseudo-labels to the unlabeled real images and used them with triplet distribution loss and conditional discriminators. The class embeddings were uniformly sampled for each conditional discriminator to mitigate the effect of imbalanced proportion of classes. 

Results with $127$ labeled target examples on test set are summarized in the main body of the paper. We use three performance metrics, namely: mean Average Precision (mAP), classification accuracy on real and sim images on a held-out test set, and Silhouette score (SS). Note that the Silhouette score (SS) is an unsupervised metric measures the tightness of a cluster relative to the other clusters without using any labels; while the classification accuracy and the mAP are supervised metrics that use the labels of the test set. We observe that the object recognition model trained on synthetic data only gives poor performance on real data with $26.8\%$ accuracy, in comparison to $85.9\%$ accuracy obtained with training on real labeled data only. Naively combining the synthetic and real data in a mini-batch is also sub-optimal. Using triplet loss on labeled source and target examples preserves the structure of the features for transfer to real examples. DANN improves performance in both domains by aligning marginal distributions. MCD further improves the performance in domain adaptation with conditional alignment of distributions. DIRL outperforms the compared approaches by combining marginal and conditional distributions alignment with triplet distribution loss. 

\begin{figure}[!tbp]
\centering
\includegraphics[trim={2.9cm 5.5cm 2cm 0.6cm},clip,scale = 0.6]{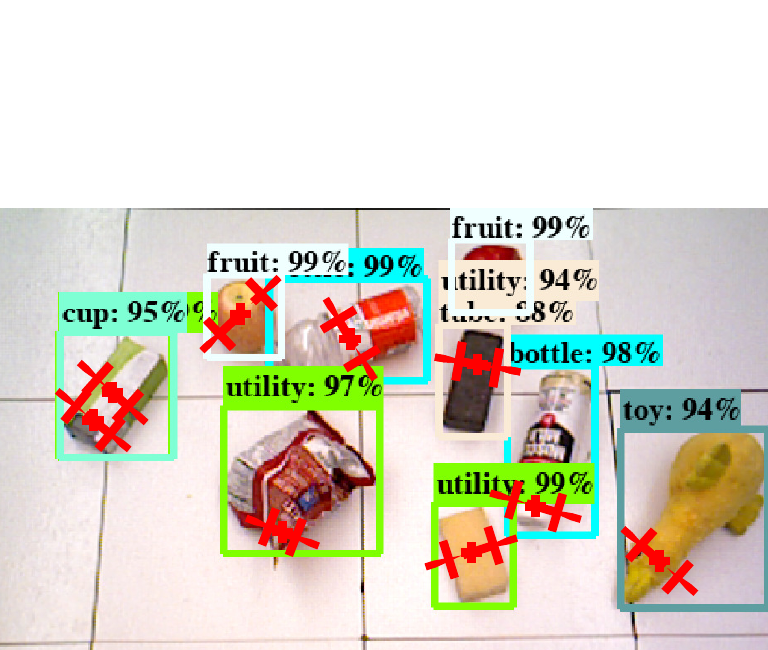}
\includegraphics[trim={3.7cm 5.5cm 2.cm 0.6cm},clip,scale = 0.6]{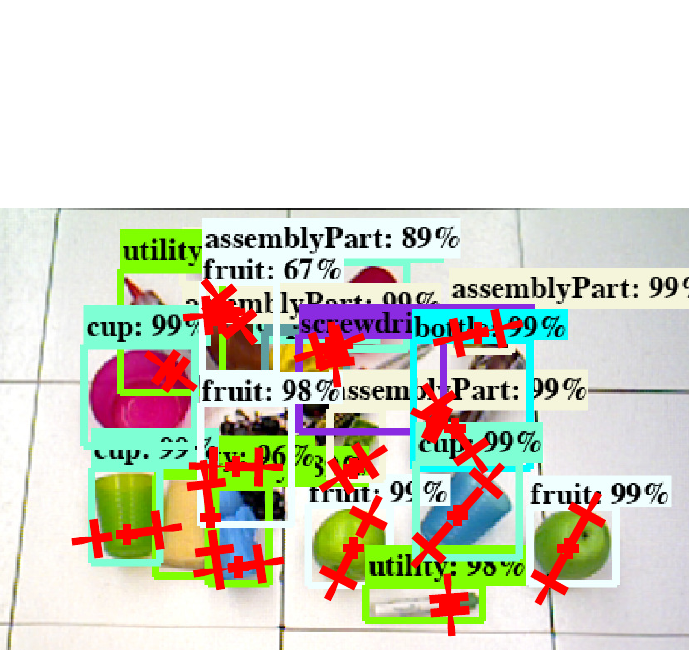}
\includegraphics[trim={1.9cm 5.5cm 2.9cm 1.2cm},clip,scale = 0.6]{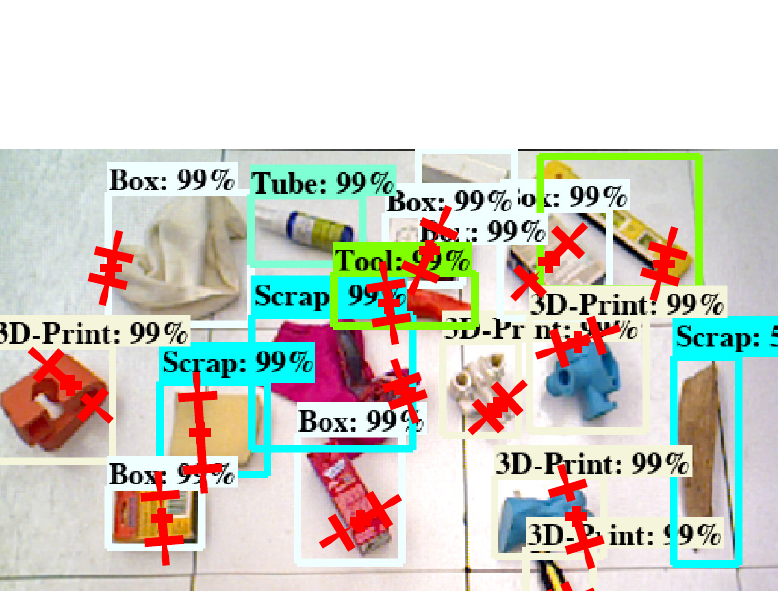}
\includegraphics[trim={4cm 8cm 6cm 4cm},clip,scale = 0.28]{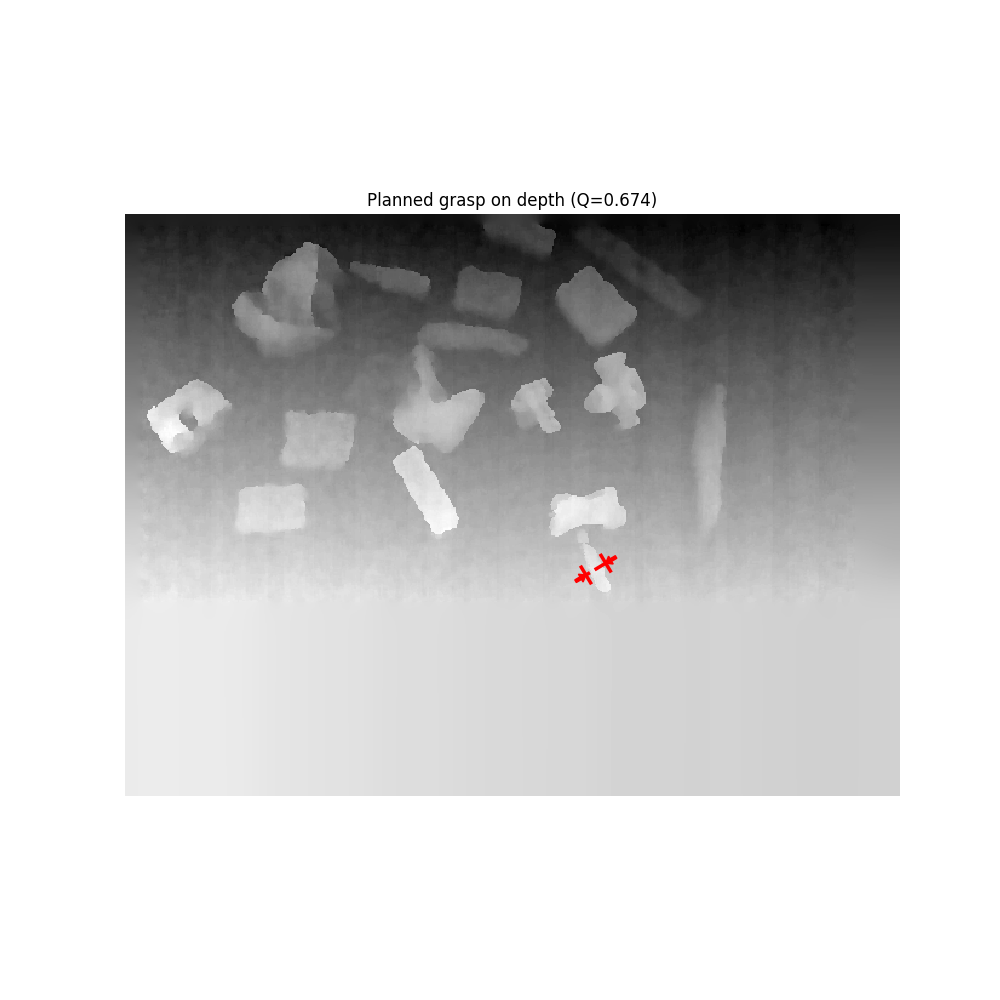}
\includegraphics[trim={4cm 8cm 6cm 4cm},clip,scale = 0.28]{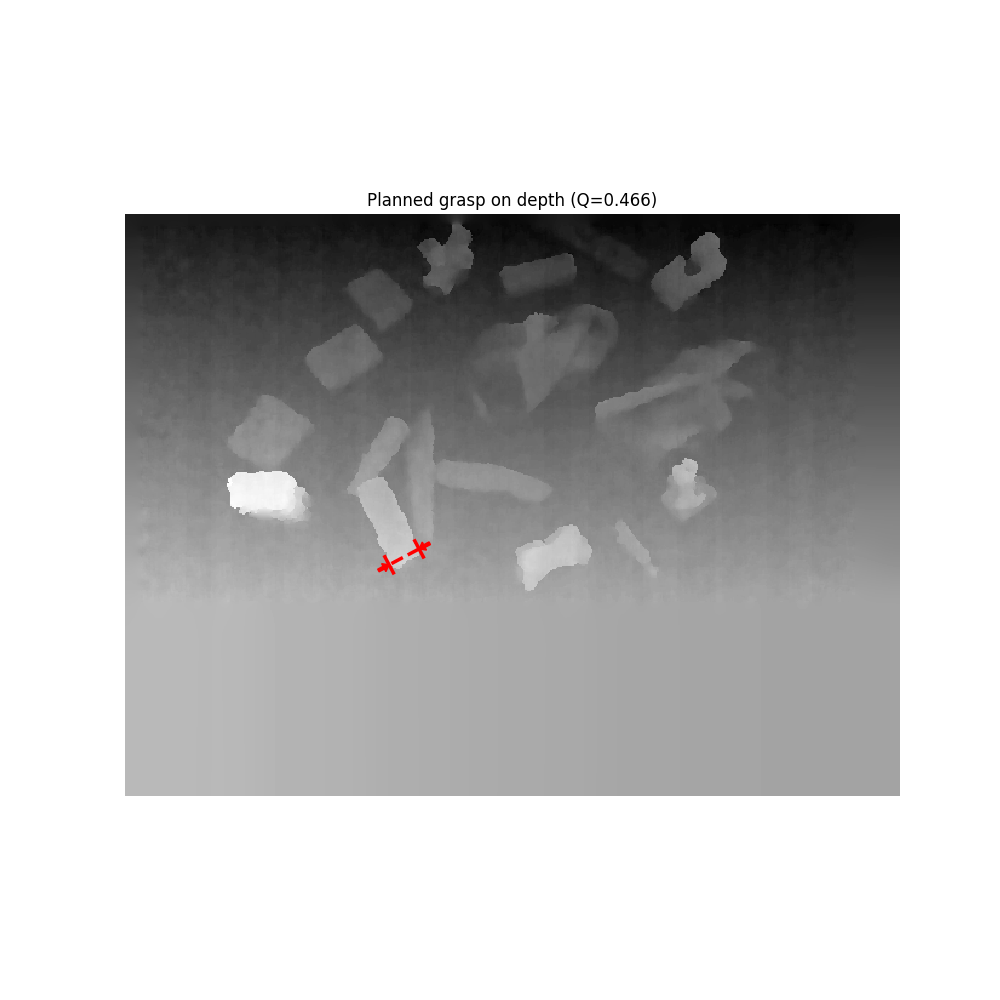}
\includegraphics[trim={4cm 6cm 4cm 7.29cm},clip,scale = 0.28]{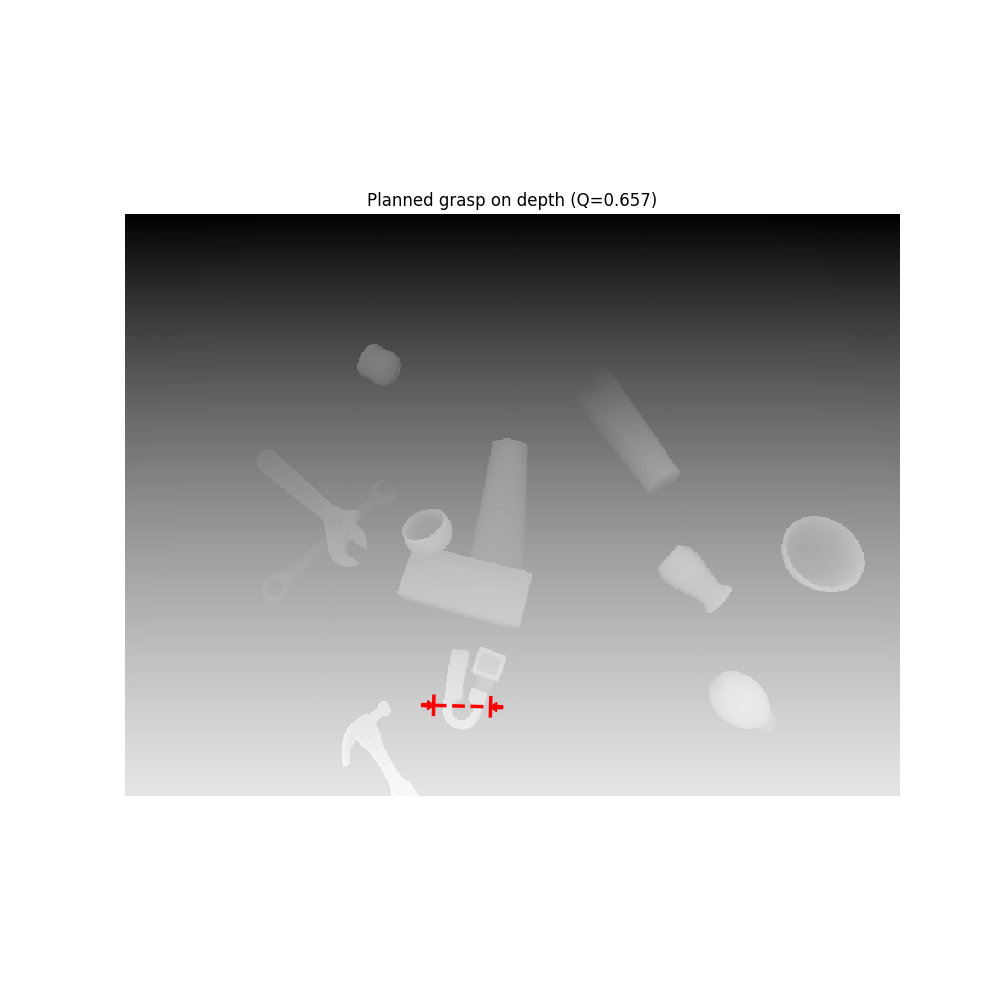}
\caption{\footnotesize Vision-Based decluttering examples of object recognition and grasp planning. \textit{(top)} Output bounding boxes from the object recognition network are fed to the grasp planning network to yield the most likely grasp to success shown with red whisker plots, \textit{(bottom)} examples of grasping a target object with real and simulated depth images.} \label{Fig: hsr_setup}
\end{figure}



\textbf{Grasp Planning Network: }The grasp planning model comprises of two parts: \textit{grasp sampling} and \textit{grasp evaluation}. The grasp planning model samples antipodal pairs on the cropped depth image of the object and  and feeds them to a convolutional neural network to predict the probability of successful grasp as determined by the wrench resistance metric. Good grasps are successively filtered with a cross-entropy method to return the most likely grasp for the robot to pick and place the object into corresponding bin.

The cropped depth image is centered around the sampled grasp center and orientation to make the network predictions rotation-invariant. The network takes a downsampled $96 \times 96$ depth image around the grasp center, and processes it with a convolution layer of size $9 \times 9$ with $16$ filters and $3$ convolution layers of size $5 \times 5$ with $16$ filters each, followed by fully connected layer of size  $128$. The average height of the grasp center region from the camera is processed through a separate fully connected layer of size $16$ before it is concatenated with the image stream features and a fully connected layer of size $128$, followed by the output layer of $2$ neurons to predict the grasp quality score $\in [0, 1]$.

We test the performance of the trained models on the Toyota HSR for picking objects from the floor and depositing them in target bins. We load $5 - 25$ objects in a bin from a set of $65$ physical objects and drop them on the floor in front of the robot. The objects may overlap after dropping; a pushing primitive is used to singulate the cluttered objects if the overlap is more than a threshold. The domain-invariant object recognition model gives $89.4 \%$ accuracy in real experiments. The grasping network gives $86.5 \%$ accuracy of picking up the identified object. Without using the grasping network and only grasping orthogonal to the principal axis of the point cloud of the predicted object location gives $76.2 \%$ accuracy. We observe that the robot performs well in grasping compliant objects and objects with well-defined geometry such as cylinders, screwdrivers, tape, cups, bottles and utilities; while assembly parts and small bowls in inverted pose induced repeated failures in grasping the target object (see video on the project website for more details).

\end{document}